\documentclass[lettersize,journal]{IEEEtran}
\usepackage{amsmath,amsfonts}
\usepackage{algorithmicx}
\usepackage{algorithm}
\usepackage[noend]{algpseudocode}
\usepackage{array}
\usepackage[caption=false,font=normalsize,labelfont=sf,textfont=sf]{subfig}
\usepackage{textcomp}
\usepackage{stfloats}
\usepackage{url}
\usepackage{verbatim}
\usepackage{graphicx}
\usepackage{cite}
\usepackage{color}
\usepackage{algorithm}
\usepackage{algpseudocode}

\newtheorem{assumption}{Assumption}

\allowdisplaybreaks

\begin{document}

\title{Conformal Predictive Safety Filter \\for RL Controllers in Dynamic Environments}

\author{Kegan J. Strawn, Nora Ayanian, and Lars Lindemann}

\maketitle

\begin{abstract}
The interest in using reinforcement learning (RL) controllers in safety-critical applications such as robot navigation around pedestrians motivates the development of additional safety mechanisms. Running RL-enabled systems among uncertain dynamic agents may result in high counts of collisions and failures to reach the goal. The system could be safer if the pre-trained RL policy was uncertainty-informed. For that reason, we propose \emph{conformal predictive safety filters} that: 1) predict the other agents’ trajectories, 2) use statistical techniques to provide uncertainty intervals around these predictions, and 3) learn an additional safety filter that closely follows the RL controller but avoids the uncertainty intervals. We use conformal prediction to learn uncertainty-informed predictive safety filters, which make no assumptions about the agents' distribution. The framework is modular and outperforms the existing controllers in simulation. We demonstrate our approach with multiple experiments in a collision avoidance gym environment and show that our approach minimizes the number of collisions without making overly-conservative predictions.
\end{abstract}


\section{Introduction}
While impressive progress has been made in deep reinforcement learning (RL) for motion planning, it remains challenging to ensure the safety of RL policies \cite{Kmmerle2013Navigation, Everett2019Collision, Chen2017Socially, Chen2017Decentralized, Everett2018Motion}. Developing safe policies for dynamical systems that operate around dynamic agents is a fundamental challenge in robotics. Other agents' intents and policies are usually unknown yet critical to any safe motion planning algorithm. An increasingly popular approach is to use RL to learn a reactive collision avoidance policy~\cite{Everett2019Collision}. However, such policies alone do not quantify uncertainty in a principled way, provide no safety guarantees, and have been observed to be unsafe. In these scenarios, when the RL policy is not guaranteed to be safe, a predictive safety filter is desirable that considers the uncertainty about the dynamic agents' future motion. Such a safety filter will ideally guarantee safety probabilistically while being minimally invasive, i.e., following the RL policy closely. Another motivation for our work is that RL policies may have been trained in different training environments or under different objectives, while a safety filter can be trained independently.

\begin{figure}[t]
  \centering
  \includegraphics[height=4cm]{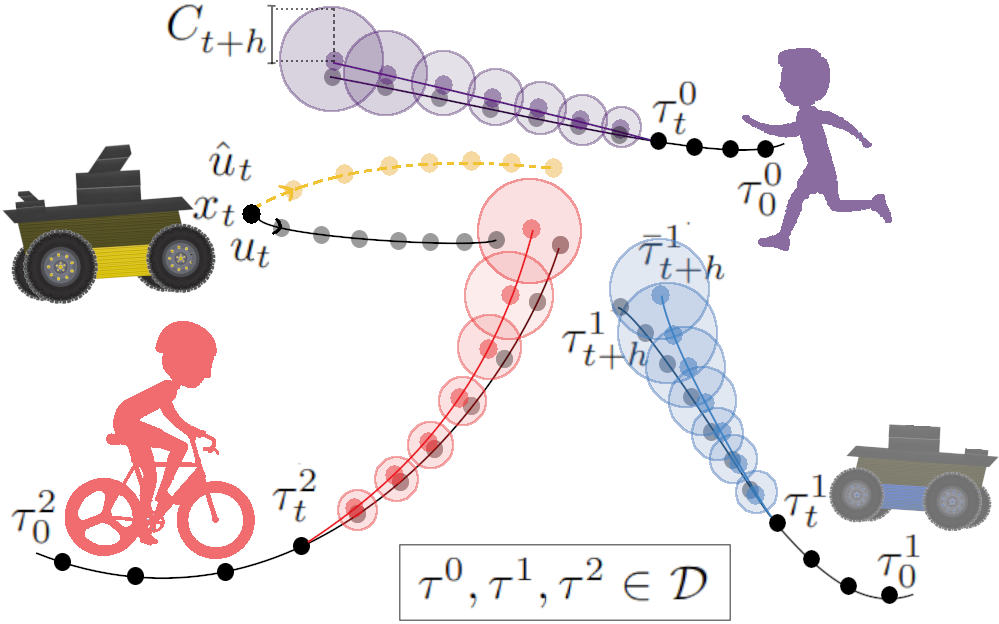}
  \caption{The system applies an RL policy $\pi$ that may lead to a collision with the other agents. Our proposed predictive safety filter $\hat{\pi}$ uses predictions of the three nearby agents (point dots) along with uncertainty intervals (shaded circles) to obtain a series of future actions that avoids collisions.}
  \label{problem_fig}
\end{figure} 

In this paper, we learn a predictive safety filter for a given RL policy that uses predicted agent trajectories, e.g., from long short-term memory networks \cite{Alahi2016Social, Hochreiter1997Long} or transformer architectures \cite{Nayakanti2022Wayformer}, along with uncertainty intervals around each prediction that are obtained using conformal prediction \cite{Angelopoulos2021Gentle}. The result is an uncertainty-informed predictive safety filter. Choosing to use conformal prediction, a distribution-free statistical tool \cite{Luo2021Sample, Dietterich2022Conformal, Bortolussi2019Neural, Fan2020Statistical, Chen2020Reactive}, we do not make any assumptions about the distribution of the agent trajectories, e.g., being Gaussian distributed. We highlight our main contributions in this paper as such: 
\begin{itemize}
\item We propose an algorithm to train uncertainty-informed predictive safety filters for pre-trained RL controllers using conformal prediction. The filter ensures probabilistic safety and incentivizes imitating the RL policy.
\item We evaluate our method in a widely-used RL collision avoidance simulator in which we reduce collisions when paired with an RL controller~\cite{Everett2019Collision} by $80\%$, reduce failures (where the agent timed out) when paired with a more conservative controller~\cite{Berg2011Reciprocal} by $67\%$, and produce shorter paths when compared to a Gaussian-based safety approach by $18\%$.
\end{itemize}

\section{Related Work}

\textbf{Planning in dynamic environments: }Model Predictive Control (MPC) is a popular planning approach that selects a minimum-cost action sequence using predictions of the agents conditioned on the current state and the history seen so far 
~\cite{Rawlings2000Tutorial}. Actions are implemented in a receding horizon fashion where only the first action is applied before new sensor measurements are obtained, and the process is repeated. Reactive-based methods use geometric or physics-based rules to ensure collision avoidance on each step and rely on a fast update rate to react to changes in the other agents' motions~\cite{Khatib1985Real, Ferrer2013Social, Berg2011Reciprocal}. 
Reactive-based RL controllers are computationally efficient but often generate sub-optimal trajectories~\cite{Chen2017Socially, Chen2017Decentralized, Everett2018Motion}. Predictive-based methods first estimate the trajectories of other agents, then plan the system's actions. These methods often yield a smoother plan but are more computationally expensive and may require additional knowledge about the other agents. It is important to note that predicted paths can be too conservative or inaccurate and block the agent's path, potentially slowing down or deadlocking all agents, known as the freezing robot problem~\cite{Trautman2015Robot}. 
Some predictive and behavior-based planners mimic what a human (or another robot) would do, require expert demonstrations, estimate the cost functions of other agents, or perform some form of additional work to understand the intents of other agents~\cite{Bojarski2016End, Kim2016Socially, Kretzschmar2016Socially, Pfeiffer2016Predicting, Tai2017Virtual, Farid2023Task, Long2017Towards}. Learning-based controllers attempt to predict the trajectories of pedestrians that could be used in collision avoidance systems or predict the reaction-based next step. An example from recent work learns from complex and multi-modal distributions of agents' motions and predictions in real-time for planning~\cite{Meszaros2023Trajflow}. Other efforts look to integrate machine learning and model predictive control under uncertainty~\cite{Mesbah2022Fusion, Wabersich2019Probabilistic}.

\textbf{Uncertainty Quantification: }We use conformal prediction (CP), see \cite{Angelopoulos2021Gentle}, for quantifying the uncertainty of trajectory predictions \cite{Lindemann2022Safe, Stankeviciute2021Conformal}. Alternative methods model the underlying distribution as a Gaussian distribution and use Kalman filters or apply  methods for finding safe sets such as forward and backward reachability~\cite{Berkenkamp2015Safe, Thrun2005Probabilistic, Niu2021Safety, Ames2021Control, Rober2023Backward, Muntwiler2020Distributed}. These alternatives can be helpful but are often overly conservative, computationally complex, or make unrealistic assumptions. Similar to CP, a Bayesian framework provides probabilistic uncertainty quantification but requires access to prior knowledge about the distribution the data is sampled from, and probably approximately correct (PAC) learning theory can be used for producing upper bounds on the probability of error for a given algorithm and confidence level, but the results often involve large constants for the overall algorithmic error~\cite{Papadopoulos2008Inductive}. 

CP provides distribution-free uncertainty quantification in such scenarios and has generally been used to quantify the uncertainty of machine learning models~\cite{Angelopoulos2021Gentle, Vovk2005Algorithmic, Shafer2008Tutorial, Fontana2023Conformal}. CP is an increasingly popular approach to obtain guarantees on a predictor's false negative rate, estimate reachable sets, and design model predictive controllers with safety guarantees~\cite{Luo2021Sample, Dietterich2022Conformal, Bortolussi2019Neural, Fan2020Statistical, Chen2020Reactive}. Recently, CP methods have been integrated with policy training for safety~\cite{Foffano2023Conformal, Taufiq2022Conformal}, time series forecasting~\cite{Stankeviciute2021Conformal} and MPC of robots in dynamic environments~\cite{Lindemann2022Safe, Dixit2022Adaptive}. Another work combines conformal prediction with reachable sets for efficient and safe MPC~\cite{Muthali2023Multiagent}. These approaches integrate CP into an MPC, which reduces the framework's modularity and cannot be directly applied to learning-based controllers.

\textbf{Safety Filters: }Most safety techniques in RL constrain the search during training updates, train with noise or adversarial agents, restrict the inputs to the policy, or attempt to learn the uncertainty of the entire system~\cite{Garcia2015Comprehensive, Berkenkamp2017Safe, Koller2018Learning, Brunke2022Safe, Wabersich2019Probabilistic}. Work exists in the safe exploration of action spaces, such as filtering unsafe actions to prevent immediate negative consequences~\cite{Dalal2018Safe}. The training safety methods have been shown to perform negligibly better than non-safe RL methods, and many robotic applications may prevent re-training, necessitating safety methods for pre-trained RL controllers~\cite{Dalal2018Safe, Glossop2022Characterising, Tearle2021Predictive, Wabersich2021Predictive}. While there could be an advantage in using CP during the training of the RL policy, we design in this paper a predictive safety filter for a pre-trained RL policy to enable the use of high-performance, off-the-shelf controllers and to increase generalizability to different applications as well as provide stronger safety guarantees.

First introduced in~\cite{Seto1998Simplex}, using safety filters for a controller in a closed-loop system has seen continued growth. Predictive safety filters assess if a proposed learning-based control input can lead to constraint violations and modify it if necessary to improve safety for future time steps. Many techniques exist, such as control barrier functions for verifying and enforcing system safety~\cite{Prajna2004Safety, Wieland2007Constructive, Ames2021Control, Wabersich2021PredictiveCB}, learning frameworks integrated with control barrier functions~\cite{Taylor2020Learning, Didier2022Approximate}, safety certification that continuously solves the optimization problem for a safe set at every online step~\cite{Didier2021Adaptive}, and model predictive control safety filters with system level synthesis have been proposed~\cite{Leeman2022Predictive}. These approaches can provide system guarantees but require explicitly modeling a system's safety requirements which are not trivial to design or implement, can be overly restrictive in the safe action sets, and increase online computational effort. 

\section{Problem Formulation: Safety Filtering}
We first define the safety filtering problem in which we would like to find a control policy (the safety filter) that closely follows a given policy (a pre-trained RL controller) while ensuring that other dynamic agents are avoided. For this purpose, consider the discrete-time dynamic control system: 
\begin{equation}\label{eq:system}
    \begin{aligned}
    x_{t+1} = f(x_{t}, u_{t})\text{, } x_{0} := \zeta
    \end{aligned}
\end{equation}
where $x_t \in X \subseteq \mathbb{R}^{N}$ and $u_t \in U \subseteq \mathbb{R}^{P}$ denote the state and the control input at time $t \in \mathbb{N} \cup \{0\}$. The sets $U$ and $X$ denote the permissible control inputs and the system's workspace. The measurable function $f: \mathbb{R}^{N} \times \mathbb{R}^{P} \rightarrow \mathbb{R}^{N}$ describes the system dynamics and $\zeta \in \mathbb{R}^{N}$ is the initial condition of the system. The system operates in an environment with $A := \{1, 2, \cdots, m\}$ dynamic agents whose trajectories are unknown as they move from start to goal locations. Specifically, let $\mathcal{D}$ be an unknown distribution over agent trajectories, and let $(\mathcal{T}_{0}, \mathcal{T}_{1}, \cdots) \sim \mathcal{D}$ describe a random trajectory where the stacked agent states $\mathcal{T}_{t} := (\mathcal{T}^{1}_{t}, \cdots , \mathcal{T}^{m}_{t})$ at time $t$ is drawn from $\mathbb{R}^{2m}$. In this paper, each agent is assumed to operate in $\mathbb{R}^{2m}$ with a two-dimensional position, but we remark that our method is not limited to $\mathbb{R}^{2}$. We use $\tau_{t} := \{\tau^{1}_{t}, \cdots, \tau^{m}_{t}\}$ when referring to a realization of $\mathcal{T}_{t}$ and assume access to the history of observations $\tau_{0:t} := \{\tau_{0}, \cdots, \tau_{t}\}$ online at time $t$. For example, the agent trajectories in Figure~\ref{problem_fig} can be described by distributions $D_{1}, D_{2}, D_{3}$ with a joint distribution $\mathcal{D}$. We make no assumptions on the form of the distribution $\mathcal{D}$ but assume i) that $\mathcal{D}$ is independent of the system \eqref{eq:system}, and ii) the availability of calibration data independently drawn from $\mathcal{D}$. We comment on these assumptions in our experiments. 

\begin{assumption} \label{interference_assumption}
For any time $t \geq 0$, the control inputs $(u_{0},\cdots,u_{t-1})$ and the resulting trajectory $(x_{0},\cdots,x_{t})$, following (\ref{eq:system}), do not change the distribution of $(\mathcal{T}_{0}, \mathcal{T}_{1}, \cdots) \sim D$.
\end{assumption}

Assumption~\ref{interference_assumption} holds approximately in many robotic applications, such as autonomous vehicles, where a car is likely to behave in ways that result in socially acceptable trajectories. In our experiments, we argue and show that the system is unlikely to drastically change the behavior of other agents, in which case conformal prediction still provides valid guarantees \cite{Cauchois2023Robust}. We later comment on such distribution shifts in more detail and reserve a thorough treatment for future work. We also assume the availability of training and calibration data drawn from $\mathcal{D}$.

\begin{assumption} \label{data_assumption}
We have a dataset of trajectories $D := \{\tau^{(0)}, \tau^{(1)}, \cdots, \tau^{(K)}\}$ in which each of the $K$ trajectories $\tau^{(i)}:= \{\tau_{0}^{(i)} , \tau_{1}^{(i)}, \cdots\}$ is independently drawn from $\mathcal{D}$.
\end{assumption}

Assumption~\ref{data_assumption} is not restrictive as large amounts of data can be obtained from rapidly advancing high-fidelity simulators or from robotic applications such as autonomous vehicles where datasets are becoming widely available. Lastly, we split the dataset $D$ into training datasets $D_{Ytrain}$ and $D_{train}$ from which we will train trajectory predictor and safety filter, respectively, and a calibration dataset $D_{cal}$ from which we will quantify the uncertainty of the trajectory predictor.
 
For this system, we are given a pre-defined controller $\pi: \mathbb{R}^{N} \times \mathbb{R}^{2m} \rightarrow \mathbb{R}^{P}$ providing the control inputs
\begin{equation}\label{eq:rl_policy}
    \begin{aligned}
    u_{t} := \pi(x_{t}, \tau_{t}).
    \end{aligned}
\end{equation} 
Here, the policy $\pi$ is an RL policy that aims to reach a final location while avoiding collisions with agents in $A$. However, RL policies may not always be successful at this task. Our goal is to construct a predictive safety filter that closely follows the RL policy while guaranteeing that the probability of a collision is upper bounded by a desired failure probability. 

\section{Conformal Predictive Safety Filter} 
\begin{figure}[t]
  \centering
  \includegraphics[width=0.85\columnwidth]{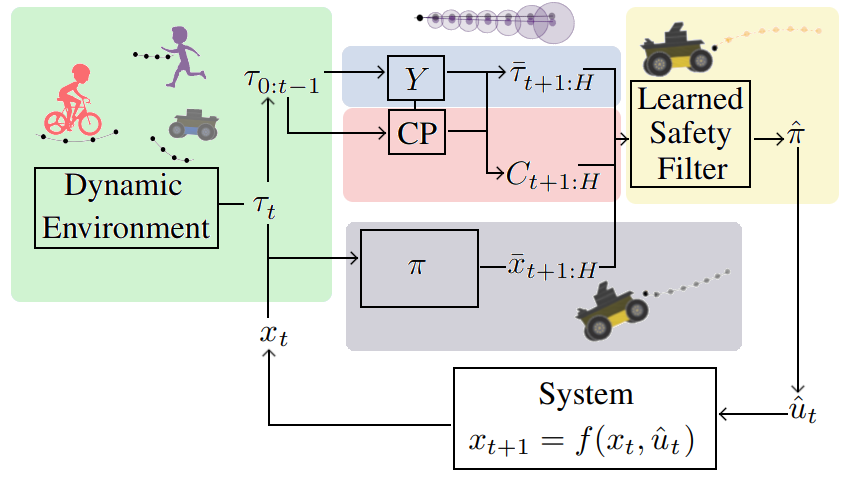}
  \caption{Overview of our predictive safety filter that produces a control policy $\hat{\pi}$ for the potentially unsafe RL policy ${\pi}$ .}
  \label{method_fig}
\end{figure}

We provide an overview of our technique upfront in Figure~\ref{method_fig}. The goal is to add a safety filter $\hat{\pi}$ to the pre-defined RL policy $\pi$ that may not have any safety certification or is only valid under certain assumptions, e.g., $\mathcal{D}$ being Gaussian. Specifically, we use a trajectory predictor $Y$ to predict future agent trajectories from past agent observations and conformal prediction (CP) to obtain uncertainty intervals for these predictions. The predictive safety filter uses this information to achieve the safety of the RL policy $\pi$ while minimally deviating from $\pi$. In the remainder, we explain the offline training of our safety filter $\hat{\pi}$, as summarized in Algorithm~\ref{alg1}. The online execution of $\hat{\pi}$ is summarized in Algorithm~\ref{alg2} and later explained and validated.

\textbf{Trajectory Predictor:} Given a prediction horizon $H$ and the history of agent observations $\tau_{0:t}$, we desire a trajectory predictor $Y: \mathbb{R}^{(t+1)2m} \rightarrow \mathbb{R}^{2mH}$ that predicts the $H$ future agent states $(\mathcal{T}_{t+1},\hdots,\mathcal{T}_{t+H})$ as $\bar{\tau}_{t+1:H} := Y(\tau_{0:t})$ where
\begin{equation}\label{eq:predictions}
    \begin{aligned}
    Y(\tau_{0:t}) := (\bar{\tau}_{t+1}, \cdots, \bar{\tau}_{t+H}).
    \end{aligned}
\end{equation} 
In principle, we can use any trajectory predictor $Y$, e.g.,  long short-term memory networks \cite{Alahi2016Social, Hochreiter1997Long} or transformer architectures \cite{Nayakanti2022Wayformer}. For training $Y$, we independently sample from $\mathcal{D}$ a dataset $D_{Ytrain}$ with trajectories from time $0$ to time $T$, i.e.,  $\tau^{(i)}_{0:T} := (\tau_{0}^{(i)}, \cdots, \tau_{t}^{(i)}, \tau_{t+1}^{(i)}, \cdots,  \tau_{T}^{(i)})$ is the $i$th trajectory in the dataset $D_{Ytrain}$. In this work, we implement our predictor by training a long short-term memory network by minimizing the following loss function over the training set $D_{Ytrain}$ (line~\ref{alg1:Ylossfunction} in Algorithm~\ref{alg1}):  
\begin{equation}\label{eq:train_pred}
    \begin{aligned}
    \min_{Y} \frac{1}{|D_{Ytrain}|}\sum_{i=1}^{|D_{Ytrain}|}\|\tau^{(i)}_{t+1:H} - Y(\tau_{0:t}^{(i)})\|^2
    \end{aligned}
\end{equation} 

\textbf{Conformal Prediction Regions:} We use CP to construct regions around the predicted trajectories that contain the true but unknown trajectory with high probability. For a general introduction to CP, we refer the reader to~\cite{Angelopoulos2021Gentle}. For trajectory predictions, the authors in~\cite{Lindemann2022Safe, Stankeviciute2021Conformal} present a technique to construct valid prediction regions applied to recurrent neural networks, which we briefly summarize next. Given observations $\tau_{0:t} := (\tau_{0}, \cdots, \tau_{t})$ at time $t$, where $\tau_{t} := (\tau^{0}_{t}, \cdots, \tau^{m}_{t})$, we can use the trajectory predictor $Y$ to obtain predictions $\bar{\tau}_{t+1:H} := (\bar{\tau}_{t+1}, \cdots, \bar{\tau}_{t+H})$ for the specified prediction horizon $H$. Given a failure probability of $\delta \in (0, 1)$, we seek values $C_{t+1:H} := (C_{t+1}, \cdots, C_{t+H})$ as prediction intervals around each prediction such that: 
\begin{equation}\label{eq:prob_gua}
    \text{Prob}(||\tau_{t+h} - \bar{\tau}_{t+h}|| \leq C_{t+h}, \text{ }\forall h \in \{1, \cdots, H\}) \geq 1 - \delta
\end{equation}
The approach is summarized in lines~\ref{alg1:CPStart} -~\ref{alg1:CPEnd} of Algorithm~\ref{alg1}. First, we define the non-conformity score function $R_{t+h} := ||\tau_{t+h} - \bar{\tau}_{t+h}||$ in line~\ref{alg1:Nonconformity}, and we evaluate the predictor based on this function across a conformal calibration dataset $D_{cal}$ that we independently sample from the distribution $\mathcal{D}$. Note that a small non-conformity score corresponds to accurate predictions, while a large score indicates that $Y$ is inaccurate. Second, we sort the non-conformity scores from the calibration dataset $D_{cal}$ in non-decreasing order and append infinity as the $(|D_{cal}| + 1)$-th value (see lines 8 and 9 in Algorithm 1). Third, we define $p := \lceil (|D_{cal}| + 1)(1 - \bar{\delta})\rceil$ (line~\ref{alg1:definep} of Algorithm~\ref{alg1}) and let the prediction interval $C_{t+h}$ correspond to the $p$th quantile over the sorted non-conformity scores for each prediction step $h\in\{1,\hdots,H\}$ (line~\ref{alg1:eachpredstep} of Algorithm~\ref{alg1}). From~\cite[Theorem 1]{Lindemann2022Safe}, we set $\bar{\delta}:=\delta/T$ in line~\ref{alg1:deltaoverT} to ensure collision avoidance with a probability of at least $1-\delta$ across the steps. Finally, when we make predictions online, we use the values $C_{t+1:H}$ as prediction intervals around each predicted trajectory $\bar{\tau}_{t+1:H}$.

\textbf{Training Predictive Safety Filters:} Our approach to training a predictive safety filter $\hat{\pi}$ is to 1) forward simulate the system from \eqref{eq:system} under the nominal RL policy $\pi$ from \eqref{eq:rl_policy} using the trajectory predictions from \eqref{eq:predictions}, and 2) enforce that the trajectory of the system from \eqref{eq:system} under the safety filter $\hat{\pi}$ imitates the trajectory under the RL policy $\pi$ while incorporating the conformal predictions regions $C_{t+1:H}$ to account for uncertainty in the trajectory predictions. 

For the first step, we use the pre-defined RL policy $\pi$ and the predictions from $Y$ to forward simulate the system dynamics under the RL 
policy into the future by $H$. We obtain this nominal future trajectory $\bar{x}_{t+1:H} := (\bar{x}_{t+1}, ..., \bar{x}_{t+H})$ as
\begin{equation}
    \begin{aligned}
    \bar{x}_t :=& x_{t} \\
    \bar{u}_{t} :=& \pi(x_{t}, \tau_{t}) \\
    \bar{x}_{t+h} :=& f(\bar{x}_{t+h-1}, \bar{u}_{t+h-1}) \text{, } \forall h \in \{1, ..., H\} \\
    \bar{u}_{t+h} :=& \pi(\bar{x}_{t+h}, \bar{\tau}_{t+h}) \text{, } \forall h \in \{1, ..., H-1\} \\
    \end{aligned}
\end{equation}

\begin{algorithm}[t]
\caption{Offline Training \& CP Calibration}\label{alg:alg1}
\begin{algorithmic}[1]
    \State \textbf{Input:} Failure probability $\delta$, horizons $H$ and $T$, pre-trained controller $\pi$, $D_{Ytrain}$, $D_{cal}$, $D_{train}$ from $\mathcal{D}$, 
    \State \textbf{Output:} Conformal Predictor $Y$, CPSF-controller $\hat{\pi}$
    \State Learn trajectory predictor $Y$ from $D_{Ytrain}$ as in \eqref{eq:train_pred} \label{alg1:Ylossfunction}
    \State $\bar{\delta} \leftarrow \delta/T$ \label{alg1:CPStart} \label{alg1:deltaoverT}
    \State $p \leftarrow \lceil(|D_{cal}| + 1)(1 - \bar{\delta})\rceil$ \label{alg1:definep}
    \For{$h$ from $1$ to $H$} 
        \State $R^{(i)}_{t+h} \leftarrow || \tau^{(i)}_{t+h} - Y({\tau}^{(i)}_{0:t+h}) ||$  \# for each $i \in D_{cal}$ \label{alg1:Nonconformity}
	\State $R^{|D_{cal}| + 1}_{t+h} \leftarrow \inf $ 
	\State Sort $R^{(i)}_{t+h}$ in non-decreasing order 
	\State $C_{t+h} \leftarrow R^{p}_{t+h}$ \label{alg1:eachpredstep}
    \EndFor \label{alg1:CPEnd}
    \For{instance $i$ in $D_{train}$} \label{alg1:trainsfstart}
        \State $\bar{\tau}^{(i)}_{t+1:H}, \leftarrow  Y(\tau^{(i)}_{0:t})$ \label{alg1:predictions_trajectories1}
        \State $\bar{x}^{(i)}_{t+1:H} \leftarrow  \text{simulate the system forward by } H \text{ using } \pi$ \label{alg1:predictions_trajectories2}
        \State $D_{sftrain}^{(i)} \leftarrow (D_{train}^{(i)}, \bar{\tau}^{(i)}_{t+1:H}, \bar{u}^{(i)}_{t:H-1}, \bar{x}^{(i)}_{t+1:H}, C_{t+1:H} )$  \label{alg1:dtrain}
    \EndFor \label{alg1:trainsfend}
    \State Learn $\hat{\pi}$, given $D_{sftrain}$ as the solution of (\ref{eq:pihat}) \label{trainSF} 
\end{algorithmic}
\label{alg1}
\end{algorithm}
Now, the safety filter $\hat{\pi}: \mathbb{R}^{2mH} \times \mathbb{R}^{HM} \times \mathbb{R}^{HN}  \times \mathbb{R}^{H}\rightarrow \mathbb{R}^{HP}$ optimizes the following objective:
\begin{equation} 
    \begin{aligned}
    \min_{\hat{\pi}} & \sum^{H}_{h=1} || \bar{x}_{t+h} - \hat{x}_{t+h} ||^{2} \label{eq:pihat}\\
    \text{s.t. }\;\;\;\hat{x}_t &:= x_{t} \\
    \hat{x}_{t+h} &:= f(\hat{x}_{t+h-1}, \hat{u}_{t:H-1}(h)) \text{, } \forall h \in \{1, ..., H\}  \\
    \hat{u}_{t:H-1}&:=\hat{\pi}(\bar{\tau}_{t+1:H}, \bar{u}_{t:H-1}, \bar{x}_{t+1:H}, C_{t+1:H})\\ 
    || \bar{\tau}^{j}_{t+h} - \hat{x}_{t+h} || &\geq C_{t+h} + \epsilon,\;  \forall h \in \{1, ..., H\},\forall j \in A
    \end{aligned}
\end{equation}
The (to be learned) safety filter produces $\hat{u}_{t:H-1}:=\hat{\pi}(\bar{\tau}_{t+1:H}, \bar{u}_{t:H-1}, \bar{x}_{t+1:H}, C_{t+1:H})$, where $\hat{u}_{t:H-1}$ contains control inputs for the next $H$ time steps. We use the notation $\hat{u}_{t:H-1}(h)$ to access the control input for the $h$th time step. The filter is recursively applied to the system across timesteps $T$. Intuitively, the safety filter minimizes the distance between the nominal RL trajectory $\bar{x}_{t+1:H}$ and the safety filter trajectory $\hat{x}_{t+1:H}$, i.e., the trajectory obtained under the safety filter $\hat{\pi}$. Additionally, the safety filter trajectory $\hat{x}_{t+1:H}$ should avoid the agent predictions $\bar{\tau}^{j}_{t+h}$ (for all agents $j\in A$) and uncertainty intervals $C_{t+h}$. Specifically, we enforce the safety constraint $||{\tau}^{j}_{t+h} - \hat{x}_{t+h} ||\ge  \epsilon$, where ${\tau}^{j}_{t+h}$ is unknown, and $\epsilon>0$ is a user-defined minimum collision avoidance distance, where $|| \bar{\tau}^{j}_{t+h} - \hat{x}_{t+h} || \geq C_{t+h} + \epsilon$ holds. Since we know $\text{Prob}(||\tau_{t+h} - \bar{\tau}_{t+h}|| \leq C_{t+h},\; \forall h\in\{1,\hdots,H\}) \geq 1 - \delta$, it is ensured that $\text{Prob}(|| {\tau}^{j}_{t+h} - \hat{x}_{t+h} || \ge \epsilon,\; \forall h\in\{1,\hdots,H\})\ge 1-\delta$.

In lines~\ref{alg1:trainsfstart}-~\ref{alg1:trainsfend} of Algorithm~\ref{alg1}, we sample another dataset $D_{train}$ of independent trajectories from $\mathcal{D}$ to train the safety filter. For each training trajectory, we construct the corresponding predictions and nominal RL trajectories (lines~\ref{alg1:predictions_trajectories1} and~\ref{alg1:predictions_trajectories2}), from which we construct the combined training dataset $D_{sftrain}$ (line~\ref{alg1:dtrain}) that we use to solve \eqref{eq:pihat}. Specifically, we train a multi-layer feedforward neural network $\hat{\pi}$ in line~\ref{trainSF}. In particular, we solve \eqref{eq:pihat} approximately over the training set $D_{sftrain}$ and rewrite the constrained optimization problem \eqref{eq:pihat} into an unconstrained optimization problem. Therefore, we augment the original cost function with the constraints of \eqref{eq:pihat} that we then minimize over to convergence. 

\begin{algorithm}[t]
\caption{Online Uncertainty Informed Safety Filter}\label{alg:alg2}
\begin{algorithmic}[1]
    \State \textbf{Input:} prediction and mission horizons $H$ and $T$, controller $\pi$, trajectory predictor $Y$, predictive safety filter $\hat{\pi}$
\For{$t$ from $0$ to $T-1$}
	\State Sense $x_{t}$ and $\tau_{t}$
	\State Compute $\bar{\tau}_{t+1:H}, \bar{u}_{t:H-1}, \bar{x}_{t+1:H}, C_{t+1:H}$ \label{compose_inputs}
	\State Compute safety filter as $\hat{u}_{t:H-1} \leftarrow \hat{\pi}(\bar{\tau}_{t+1:H}, \bar{u}_{t+1:H}, \bar{x}_{t+1:H}, C_{t+1:H})$ \label{sample_action}
        \State Apply $\hat{u}_{t:H-1}(1)$ to the system \label{alg2:apply}
    \EndFor
\end{algorithmic}
\label{alg2}
\end{algorithm}

\textbf{Online Execution: }Once the safety filter is trained, it can be used during runtime by first computing the set of $H$ next control inputs $\hat{u}_{t:H-1}:=\hat{\pi}(\bar{\tau}_{t+1:H}, \bar{u}_{t:H-1}, \bar{x}_{t+1:H}, C_{t+1:H})$. Specifically, we apply the safety filter in a receding horizon fashion by applying the first element $\hat{u}_{t:H-1}(1)$ at each time step $t$. The online execution is summarized in Algorithm~\ref{alg:alg2}. In line~\ref{compose_inputs}, we pass the history of trajectories $\tau_{0:t}$ to our predictor $Y$ to obtain predictions $\bar{\tau}_{t+1:H}$, we also obtain the nominal RL control inputs and trajectory $\bar{u}_{t:H-1}$ and $ \bar{x}_{t+1:H}$, as well as the conformal uncertainty intervals $C_{t+1:H}$. In lines~\ref{sample_action} and~\ref{alg2:apply}, we compute the safety filtered control input and apply $\hat{u}_{t:H-1}(1)$ to the system. 

\textbf{Guarantees: } Under the assumption that the safety filter achieves the objective in \eqref{eq:pihat}, we remark that our safety filter guarantees probabilistic safety, i.e., that $\text{Prob}(|| {\tau}^{j}_{t} - \hat{x}_{t} || \ge \epsilon,\; \forall t\in\{1,\hdots,H\})\ge 1-\delta$ as we use $\bar{\delta} := \delta/T$. This follows the same reasoning as in \cite[Theorems 1 and 3]{Lindemann2022Safe}. Note that these guarantees hold under idealized assumptions.
In practice, there may be reasons why we do not achieve exact probabilistic coverage. One reason is the way we approximately solve the optimization problem \eqref{eq:pihat} over the training set $D_{sftrain}$ by rewriting \eqref{eq:pihat} as an unconstrained optimization problem. In our experimental results, however, we observe that a safety filter $\hat{\pi}$ trained with sufficient data performs well in practice and that our method provides an added layer of safety over the baseline RL policy.

Furthermore, we note that the prediction intervals $C_{t+h}$ naturally depend on the underlying distribution $\mathcal{D}$, the predictor's accuracy, and the user-specified risk tolerance $\delta$. Without the intervals $C_{t+h}$, the safety filter $\hat{\pi}$ would only mimic the baseline RL controller $\pi$. Therefore, the safety filter $\hat{\pi}$ may perform differently than $\pi$. Additionally, Figure~\ref{method_fig} demonstrates the modularity of our approach. This attribute enables using potentially different policies with the learned safety filter, which we will investigate in future work. 

As per Assumption \ref{interference_assumption}, we assume that the distribution of trajectories does not change from offline training to online testing. In practice, we argue that small distribution shifts in $\mathcal{D}$ will not significantly affect our algorithm and its guarantees. A supporting argument of our claim is given in~\cite{Cauchois2023Robust}, where it is formally shown that small distribution shifts only lead to small deviations from the desired conformal prediction guarantees. In our experiments, we explicitly check that agent interaction does not introduce a larger distribution shift, as we will verify in Figure~\ref{dist_shift}. One may increase the robustness of the prediction intervals using adaptive conformal prediction~\cite{Dixit2022Adaptive}. 

\begin{figure}[t]
  \centering
  \includegraphics[width=\columnwidth]{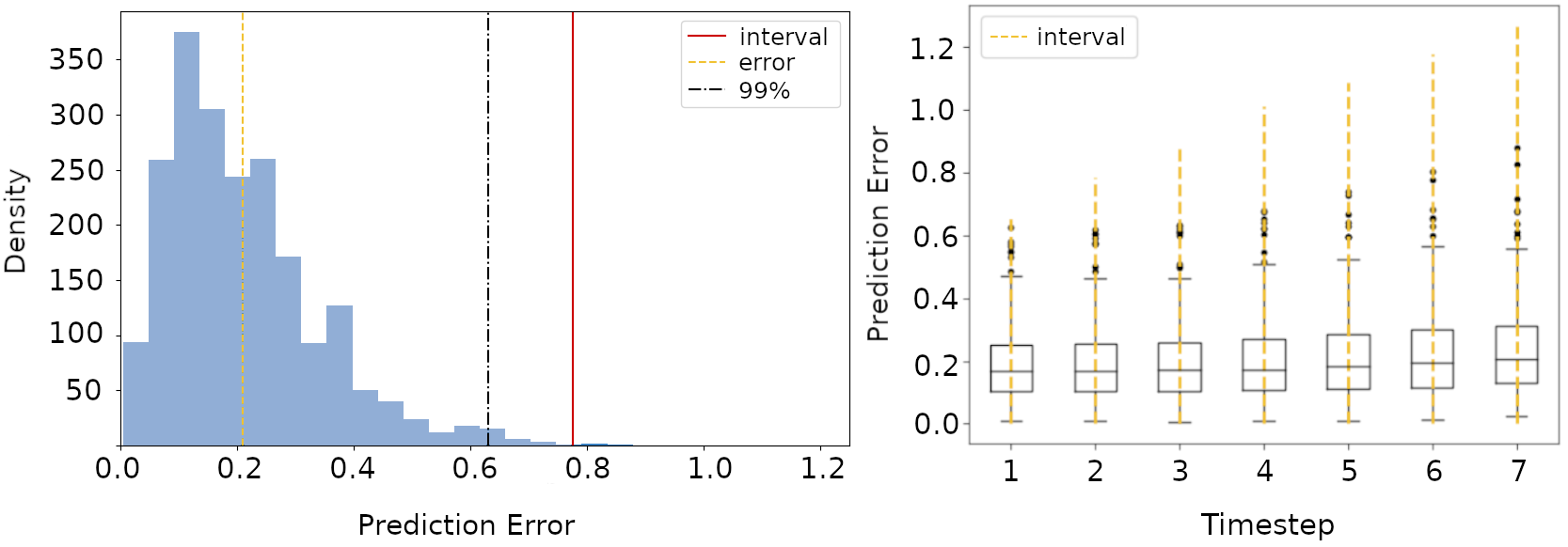}
  \caption{(Left) The average prediction error, with the conformal interval at a timestep. (Right) Boxplots of the prediction errors by timesteps ahead, with the conformal interval coverage displayed as a yellow vertical dashed line. 
  } 
  \label{conformal_figures}
\end{figure}

\section{Experimental Evaluation}\label{Experiments}
The experiments were run on a 5.0 GHz Turbo Intel Core i9-9900K computer with 16GB RAM in the Collision Avoidance Gym~\cite{Everett2019Collision}. The multiagent gym environment is open-sourced and provides pre-trained RL policies for navigation from given start to goal locations. Agents can sense the locations and velocities of the other agents. The gym allows customizable dynamic models and supports quick benchmark testing against future techniques. In \cite{Everett2019Collision}, the authors demonstrate the transferability of policies from training in their gym simulator to deployment on real-world aerial and ground robots in real-time with diverse sensors. For the dataset generation and online experiments, we consider bicycle system dynamics, a planning frequency of $10$Hz, and an agent radius of $\epsilon:=0.5$. All dynamic agents use $CADRL$, a collision avoidance RL policy trained to mimic socially acceptable pedestrian movement~\cite{Chen2017Socially}. We consider three scenarios with sets of $(2, 4, 6)$ dynamic agents $A$. For each scenario, we ran $20000$ instances of trajectories sampled from $\mathcal{D}$ that we split equally between the predictor and safety filter training datasets $D_{Ytrain}$ and $D_{train}$, respectively. An additional $1000$ trajectories were collected for the conformal calibration dataset $D_{cal}$. 

We use GA3C-CADRL (GA3C) as our given RL policy $\pi$ around agents from $\mathcal{D}$ to train our safety filter. A pre-trained policy for GA3C is available in the gym environment, has received significant attention, and has demonstrated good performance when transferred to a fully autonomous robot moving around pedestrians. The authors note that collisions still occur. For comparison, we generate a dataset and train a second safety filter that uses the more conservative reciprocal velocity avoidance algorithm ORCA controller~\cite{Fiorini1998Motion}. The flexibility of our method allows us to apply our safety filter to both GA3C and ORCA controllers to produce CPSF-GA3C and CPSF-ORCA. 

For uncertainty quantification of our predictor, we set $\delta := 0.01$ to provide our safety filter $99\%$ confidence during training. This can be tuned to produce larger avoidance actions or smaller ones, depending on the level of conservatism desired. The average length of an experiment is around $8$ seconds, i.e., $T:=80$, and we set $H := 7$. We tested various lengths of prediction horizons $H$ and found negligible improvement beyond $H:=7$ future steps. We set the architecture for the predictor LSTM to be $2$ layers of $128$ hidden units and the safety filter network to be $3$ layers of $128$. 

\begin{figure}[t]
  \centering
  \includegraphics[width=0.95\columnwidth]{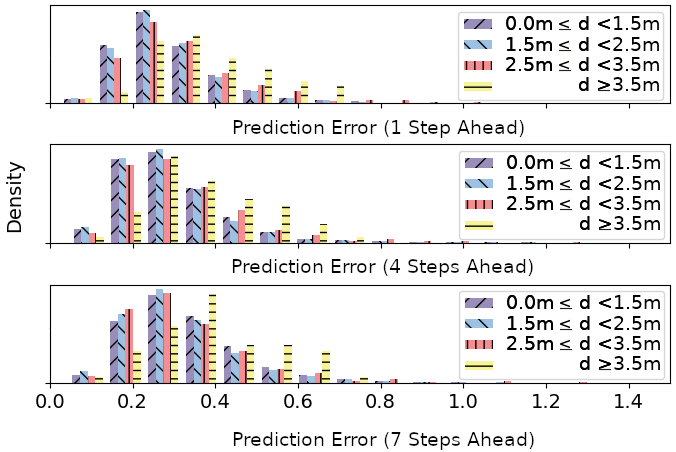}
  \caption{For a set of test trajectories, we categorize each trajectory based on the minimum inter-agent distance $d$ between all agents. We consider four different categories of minimum inter-agent distances $[0,1.5), [1.5,2.5), [2.5,3.5), $ and $[3.5,\infty)$. We use this as a proxy for analyzing agent interactions. We plot the histograms of prediction errors for each category's $1$st, $4$th, and $7$th step-ahead predictions. We can see that the histograms between the first three categories look almost identical, indicating no significant distribution shift. For the last category, where we consider larger inter-agent distances, we observe slightly different histograms.
  }
  \label{dist_shift}
\end{figure}

In Figure~\ref{conformal_figures}, we visualize our CP method performance across the calibration dataset and observe the average interval relative to all non-conformal scores. On the right, we see the  growth of the intervals with each step into the future. The environment consists of only the system and the agents sampled from the distribution $\mathcal{D}$. The agents may approach each other, and interactions in close proximity could produce a change in the distribution $\mathcal{D}$. However, to verify that Assumption \ref{interference_assumption} approximately holds, we group trajectories from a test set by their minimum inter-agent distances of all agents into four categories $[0,1.5), [1.5,2.5), [2.5,3.5), $ and $[3.5,\infty)$.  We then plot the histograms of the prediction errors for the $1$, $4$, and $7$ step ahead prediction for each category in Figure~\ref{dist_shift}. We can see that the histograms between the first three categories look nearly identical, indicating no significant distribution shift. For the last category, the histograms show differences. However, we note that there were fewer trajectories in this category. Therefore, the histogram for that category may not be as accurate, and we note that trajectories from this category are less likely to occur in our experiments. The plot approximately justifies the independence assumption of the trajectories made in CP.

Our safety filter method is not limited to CP for statistical guarantees. By swapping out the CP for a standard Gaussian process, which learns a mean and standard deviation around each prediction as a safety guarantee, we can train a second safety filter, GASF-GA3C. 

\section{Comparative Results for Predictive Safety Filtering}\label{Results}
 
\begin{figure}[t]
  \centering
  \includegraphics[width=0.95\columnwidth]{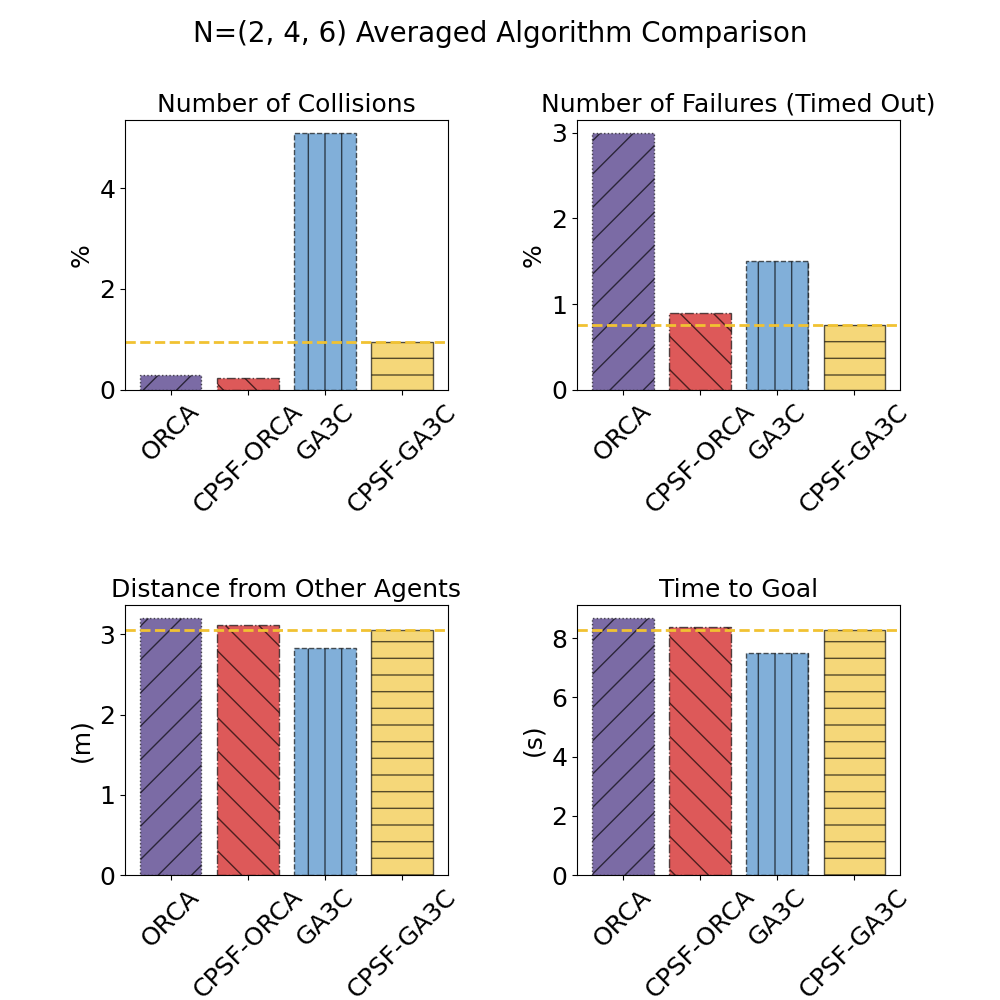}
  \caption{The averaged performance across all agent set sizes.}
  \label{results_plot}
\end{figure}

For the online experiments, we ran $1000$ instances for each set of agents $(2, 4, 6)$. We present the average performance across all agent test sets, comparing the percentage of collisions and failures (where the agent timed out), the average minimum distances to other agents, and the time to goal in Figure~\ref{results_plot}. Table~\ref{table1} lists the average performance across all instances for the $4$ agent case.

GA3C performs less conservative trajectories that lead to a shorter time to goal but a higher number of collisions than ORCA. However, ORCA performs overly conservative trajectories that reduce collisions but increase failure to reach the goal. Our conformal predictive safety filter minimizes the negative side effects of each control policy, with fewer collisions for CPSF-GA3C and fewer failures for CPSF-ORCA. We highlight that the results match our risk tolerance set by $\delta := 0.01$, with the nonzero collisions under a configurable threshold. 
GA3C, without the safety filter, has the shortest distance from other agents and time to goal, but our safety filter performs as well as the other algorithms in time to goal and minimally impacts GA3C. 

We acknowledge that the simulated gym environment is dense, and the distances between agents can be small, with subtle changes during key interactions. Fine-tuning the hyperparameters or $\delta$ could produce more drastic changes as desired. 
We note that the learned CPSF algorithms, trained to mimic the original policies, may differ from the expert policy due to conformal prediction constraints during learning, imperfect training, and finite training data.
Overall, the CPSF algorithms outperform the non-CPSF algorithms in the number of collisions and failures. The CPSF algorithms maintain similar time to goals and distances from other agents and show an awareness of the uncertainty in the underlying approaches.

\begin{table*}[htbp]
  \caption{N=4 Collision Avoidance Experiments}
  \label{table1}
  \centering
  \begin{tabular}{|l|c|c|c|c|c|}
    \hline
    \textbf{System Controller} & \textbf{Collisions} & \textbf{Failures (timed out)}  & \textbf{Distance from Other Agents (m)} & \textbf{Time to Goal (s)}  \\
    \hline
    ORCA & 3 & 27 & 3.201 & 8.678 \\
    \hline
    CPSF-ORCA & 2 & 15 & 3.131 & 8.412 \\
    \hline
    GA3C & 49 & 14 & 2.835 & 7.503 \\
    \hline 
    \textbf{CPSF-GA3C} & 10 & 8.0 & 3.057 & 8.278 \\
    \hline
  \end{tabular}
\end{table*}

\begin{table*}[htbp]
  \centering
  \caption{N=4 Safety Coverage}
  \label{table2}
  \begin{tabular}{|c|c|c|c|}
    \hline
    \textbf{Method} & \% \textbf{Inside Interval} & \textbf{Distance from Other Agents (m)}  & \textbf{Time to Goal (s)} \\
    \hline
    \textbf{CPSF-GA3C} & 99\% & 3.057 & 8.278\\
    \hline
    GASF-GA3C & 100\% & 3.860 & 10.05\\
    \hline
  \end{tabular}
\end{table*}

We ran an additional $1000$ instances with CPSF-GA3C and listed the results in Table~\ref{table2} for $4$ agents. The Gaussian approach captures the prediction uncertainty but increases the agent's time to the goal. CP does not assume a Gaussian distribution and can deliver a specified-risk level guarantee that the predictions contain the true location of the agent, resulting in a lower time to the goal. 

\begin{figure*}[htbp]
  \centering
  \includegraphics[width=0.60\columnwidth]{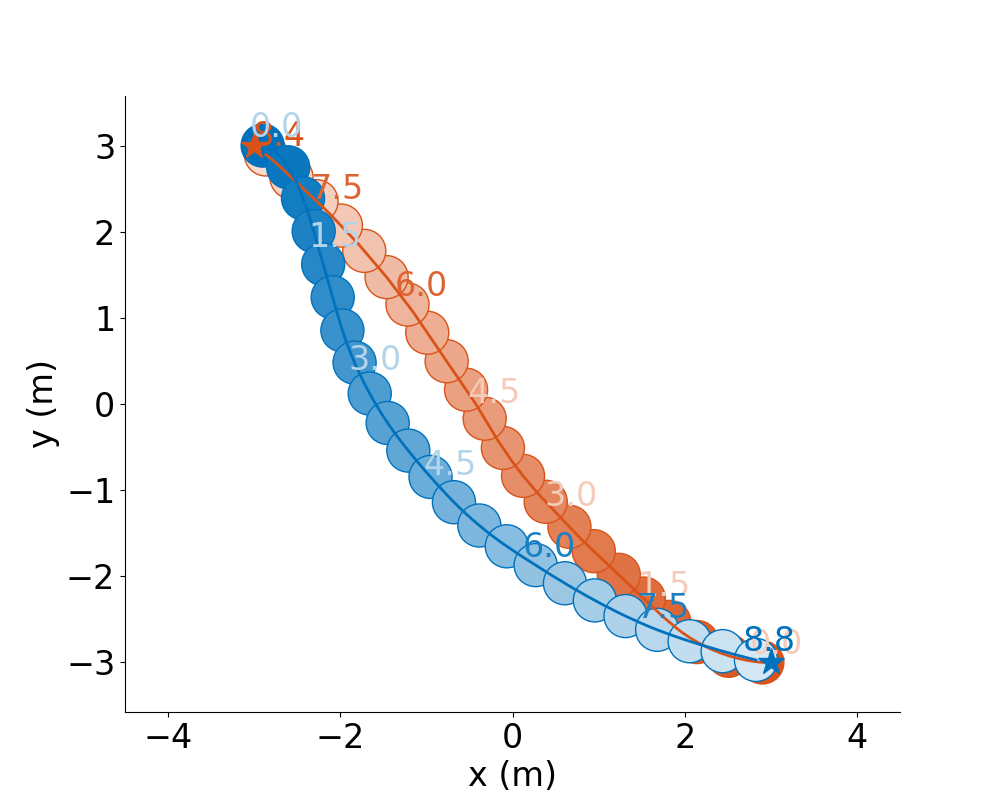}
  \includegraphics[width=0.60\columnwidth]{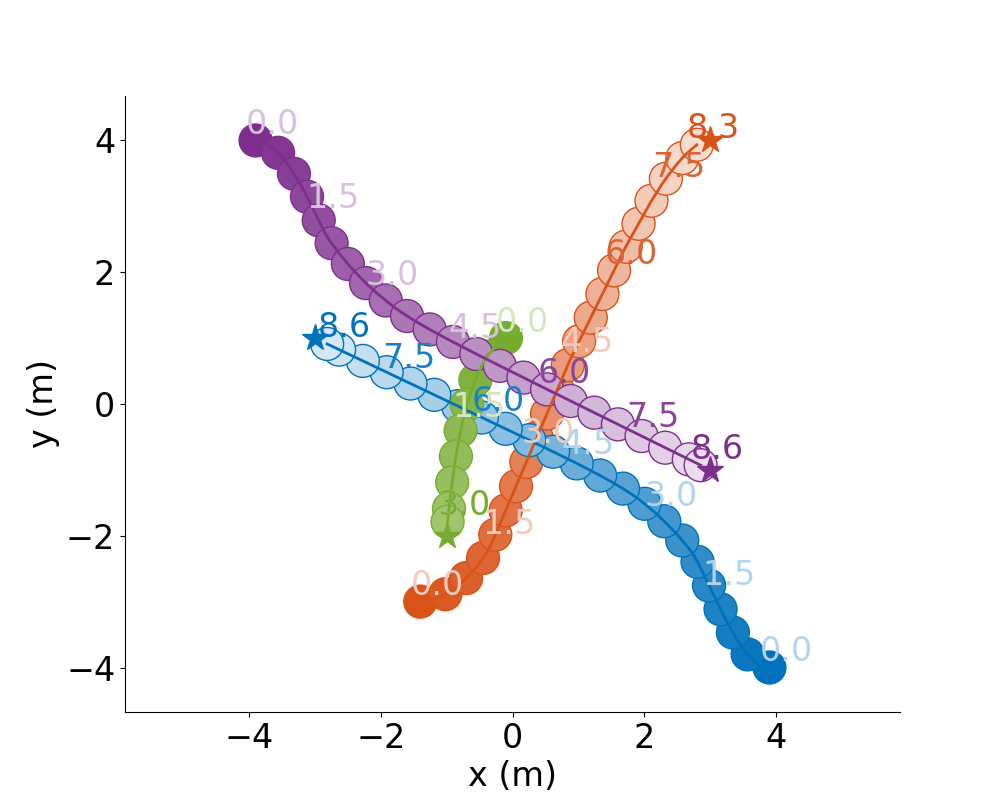}
  \includegraphics[width=0.60\columnwidth]{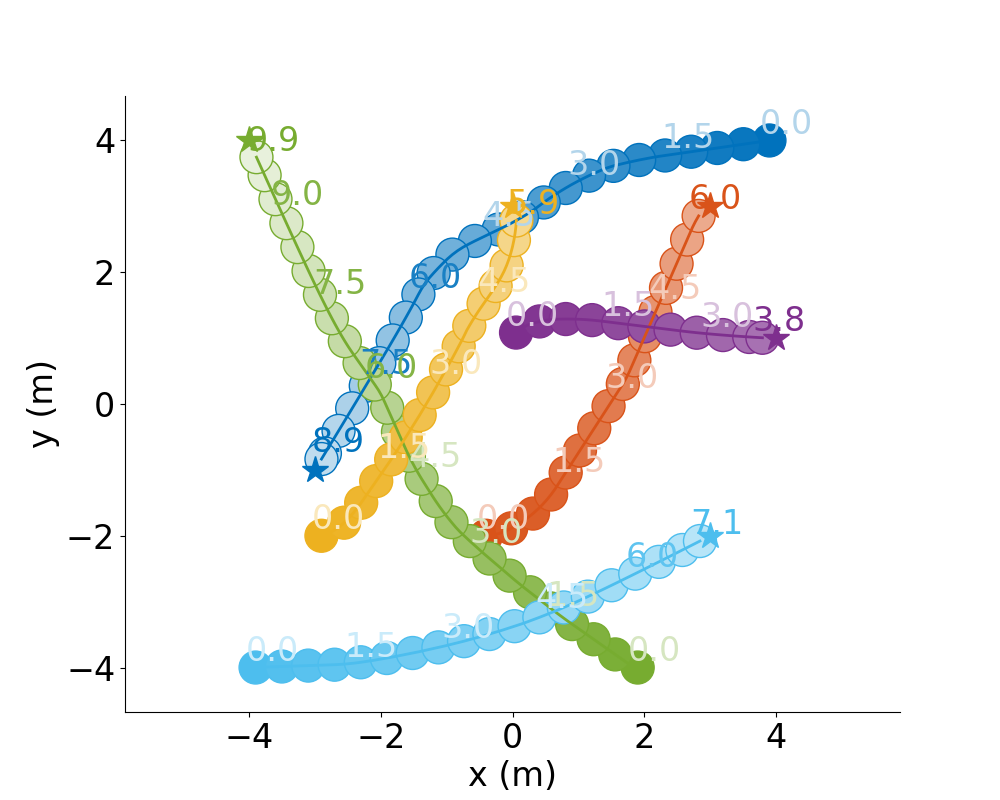}
  \caption{Our system (orange) starts at ($-3$, $3$), ($-2$, $-3$), and ($-1$, $-2$) in each case, respectively. The path darkens as time increases. 
  } 
  \label{gym_figure}
\end{figure*}

Figure~\ref{gym_figure} visualizes the trajectories, showing a run of the simulation in the collision avoidance gym. The agents perform smooth trajectories that avoid collisions while efficiently reaching their goals. Our method in a widely-used RL collision avoidance simulator achieves $80\%$ fewer collisions when paired with GA3C-CADRL and $67\%$ fewer failures when paired with ORCA, while planning paths shorter by $18\%$ than the Gaussian-based safety filter. 

\section{Conclusion}
While there has been impressive progress in RL for robot motion planning in dynamic environments, RL policies are often unsafe, and the rate of collisions can be at an undesirable level. We present a framework for learning predictive safety filters for safe motion planning by predicting future trajectories of other agents and quantifying prediction uncertainty using conformal prediction. We evaluated our framework for multiple dynamic agents in a collision avoidance gym environment that has been shown to transfer to physical robots. Our proposed predictive safety filter achieved fewer collisions without increasing the time-to-goal or number of failures to reach the goal.

In the future, we will test the predictive safety filter for other controllers (not only RL controllers) and verify the modularity, e.g., using the safety filter for a controller not considered during training. Future work on applying adaptive CP is a promising direction to relax some of the assumptions made in this work, e.g.,  when distribution shifts occur in $\mathcal{D}$.

\bibliographystyle{IEEEtran}
\bibliography{references}

\newpage
\vspace{11pt}
\vfill
\end{document}